\documentclass{article}

\usepackage{PRIMEarxiv}

\usepackage[utf8]{inputenc} % allow utf-8 input
\usepackage[T1]{fontenc}    % use 8-bit T1 fonts
\usepackage{hyperref}       % hyperlinks
\usepackage{url}            % simple URL typesetting
\usepackage{booktabs}       % professional-quality tables
\usepackage{amsfonts}       % blackboard math symbols
\usepackage{nicefrac}       % compact symbols for 1/2, etc.
\usepackage{microtype}      % microtypography
\usepackage{lipsum}
\usepackage{fancyhdr}       % header
\usepackage{graphicx}       % graphics
\usepackage{multirow}
\usepackage{xcolor}
\usepackage{soul}
\usepackage{amsmath}
\usepackage{algorithm}
\usepackage{algpseudocode}
\usepackage{hyperref}

\graphicspath{{media/}}     % organize your images and other figures under media/ folder

\title{NOTE: Notable generation Of patient Text summaries through Efficient approach based on direct preference optimization
\thanks{\textit{\underline{Citation}}: 
\textbf{Authors. Title. Pages.... DOI:000000/11111.}} 
}

\author{Imjin Ahn \\
INMED DATA \\
Yonsei University \\
Seoul \\
Republic of Korea \\
  \texttt{seraphina311@inmed-data.com} \\
    \texttt{seraphina311@yonsei.ac.kr} \\
     \And
  Hansle Gwon \\
INMED DATA\\ 
Yonsei University \\
Seoul \\
Republic of Korea \\
  \texttt{khs1220@inmed-data.com} \\
  \texttt{qwehgf1231@yonsei.ac.kr} \\
       \And
  Young-Hak Kim \\
  INMED DATA \\
  Asan Medical Center \\ 
Seoul \\
Republic of Korea \\
  \texttt{yhkim@inmed-data.com}\\
  \texttt{mdyhkim@amc.seoul.kr } \\
       \And
  Tae Joon Jun \\
  INMED DATA \\
  Asan Medical Center \\ 
Seoul \\
Republic of Korea \\
  \texttt{taejoon@inmed-data.com} \\
  \texttt{taejoon@chilab.kr} \\
       \And
  Sanghyun Park \\
Yonsei University \\
Seoul \\
Republic of Korea \\
  \texttt{sanghyun@yonsei.ac.kr} \\
}

\begin{document}
\maketitle

\begin{abstract}
The discharge summary (DS) is a crucial document in the patient journey, as it encompasses all events from multiple visits, medications, varied imaging/laboratory tests, surgery/procedures, and admissions/discharge. Providing a summary of the patient’s progress is crucial, as it significantly influences future care and planning. Consequently, clinicians face the laborious and resource-intensive task of manually collecting, organizing, and combining all the necessary data for a DS. Therefore, we propose NOTE, which stands for “Notable generation Of patient Text summaries through an Efficient approach based on direct preference optimization (DPO)”. NOTE is based on MIMIC-III and summarizes a single hospitalization of a patient. Patient events are sequentially combined and used to generate a DS for each hospitalization. To demonstrate the practical application of the developed NOTE, we provide a web page-based demonstration software. In the future, we will aim to deploy the software available for actual use by clinicians in hospital.
NOTE can be utilized to generate various summaries not only discharge summaries but also throughout a patient's journey, thereby alleviating the labor-intensive workload of clinicians and aiming for increased efficiency.

\end{abstract}

% keywords can be removed
\keywords{Text summarization\and Text generation\and Electronic medical records\and discharge summary\and Large language model\and Optimization}

\section{Introduction}
The discharge summary (DS) is a crucial document in the patient journey\cite{patient_journey}, as it encompasses all events from multiple visits, medications, varied imaging/laboratory tests, surgery/procedures, and admissions/discharge. This comprehensive data, including tables, text, imaging, and signals, is meticulously stored in the electronic medical records (EMR) system\cite{emr}. Admission and discharge represent primary steps in the patient journey. Clinicians write an initial admission report on the day of hospitalization based on the patient's history, along with a daily progress report and a DS. Especially,the DS is essential for follow-up care and treatment planning as it covers all events that occurred during the hospitalization\cite{discharge_important}.

Despite being laborious and resource-intensive, clinicians efficiently collect, organize, and combine all necessary data for the DS. It is clear that this process requires a significant amount of time and effort, and constitutes a substantial portion of the clinicians' workload. To overcome this situation, researches of artificial intelligence (AI) algorithms in medicine has been extensively conducted to enhance the efficiency and quality of medical services\cite{mimic_prediction}.

Natural language processing (NLP) stands out as a vital AI algorithm, fueling extensive research and technological advancements, particularly in the realm of large language models (LLMs)\cite{llm_medicine, llm_gpt}. These models are proving invaluable in tasks such as summarizing patient information and generating concise summaries.

In the medical domain, where texts often feature a blend of lots of languages, abbreviations, and special characters, each clinician adopts a unique writing style. Recognizing this complexity, we introduce NOTE - an acronym for "Notable generation Of patient Text summaries through an Efficient approach based on direct preference optimization (DPO)\cite{DPO}." Leveraging the Medical Information Mart for Intensive Care-III (MIMIC-III) dataset\cite{mimic_1, mimic_2, mimic_3}, NOTE excels in summarizing a patient's hospitalization journey.
By extracting and condensing crucial information from diverse medical data, NOTE streamlines the patient journey. Moreover, it has been tailored to operate effectively within hospital environments, ensuring compliance with stringent personal information protection measures and optimizing the utilization of computing resources.

The contributions of NOTE are outlined as follows:
Firstly, rather than relying solely on a single table within Electronic Medical Records (EMRs), we utilized multiple tables and textual data to construct a sequential dataset sorted by time. This dataset serves a dual purpose: not only does it facilitate model training, but it also generates individual patient reports. This approach offers clinicians the advantage of accessing comprehensive patient information in a single view.
Secondly, we introduced a method to effectively fine-tune LLMs, ensuring their optimal utilization within medical institutions burdened with extensive data and limited computing resources.
Lastly, we developed web-based demo application for NOTE, facilitating its practical implementation. This software seamlessly integrates with EMRs, enabling automated summary generation. This versatility extends beyond diagnosis support, encompassing various forms of medical text generation such as reports and progress updates.

\section{Related works}
These days, with the development of LLMs, we can see a lot of movement to apply NLP across industries. Among the various parts of NLP, the field of generative AI\cite{genAI} is the the most topical these days, and research on developing new foundation models or fine-tuning LLMs to suit downstream tasks is continuously increasing. In particular, IT companies are building extensive datasets and announcing foundation models to develop their own LLMs, and research on fine-tuning the models is also actively underway\cite{gemini, llama2}.

Summarization, one of the NLP task, was traditionally categorized into extractive and abstractive approaches\cite{extractive, abstractive}. However, the advent of text-to-text based models like T5\cite{t5} led to state-of-the-art (SOTA) results. Moreover, with the emergence of ChatGPT\cite{openai_chatgpt} and GPT4\cite{openai_assistant}, there's a growing trend towards leveraging foundation models or application programming interfaces (API). However, developing LLMs is challenging due to extensive computing requirements, leading to a preference for fine-tuned models or domain-specific API.

In the medical domain, there's significant interest in integrating AI. Most research had focused on disease or prognosis prediction using internal hospital data or datasets like MIMIC-III\cite{mimic_xgb,mimic_nature}. Furthermore, research utilizing MIMIC-III and LLMs for summary generation is in progress, but it primarily adopts the existing report format \cite{sum_progress}.

In recent years, there's been a shift towards leveraging LLMs to gather data through collaborations between hospitals and companies\cite{medpalm_google, gatortron_nvidia}. However, medical institutions strictly regulate data import/export, allowing access to de-identified data only through rigorous approval processes for research purposes. Therefore, there's a growing need for internal models or models that can be efficiently fine-tuned with minimal resources\cite{small_llm}.
NOTE introduces a novel approach by generating suitable datasets for training and employing DPO and Parameter-Efficient Fine Tuning (PEFT)\cite{peft} to train LLM.

\section{Dataset}
We collected our data from MIMIC-III\cite{mimic_1, mimic_2, mimic_3}, a comprehensive and freely accessible de-identified database. Access to this database requires a number of steps to obtain permission. MIMIC-III includes data from over forty thousand patients admitted to the intensive care units (ICU) at Beth Israel Deaconess Medical Center between 2001 and 2012. Similar to other EMRs, MIMIC-III contains a wide range of information, including demographics, vital signs, laboratory test results, procedures, medications, clinical notes, and imaging test results. 

It is structured as a relational database consisting of 26 tables, with the primary identifiers being SUBJECT\_ID (a unique patient number) and HADM\_ID (a unique hospital admission number), which link all the tables together. From these 26 tables, we extracted 12 tables that contain events that may occur during a hospitalization in the DS. These tables include PATIENTS, ADMISSIONS, DIAGNOSES\_ICD, PROCEDURES\_ICD, PRESCRIPTIONS, CHARTEVENTS, LABEVENTS, and NOTEEVENTS.

\section{Methologies}
\begin{figure*}
  \centering
  \includegraphics[width=0.8\linewidth]{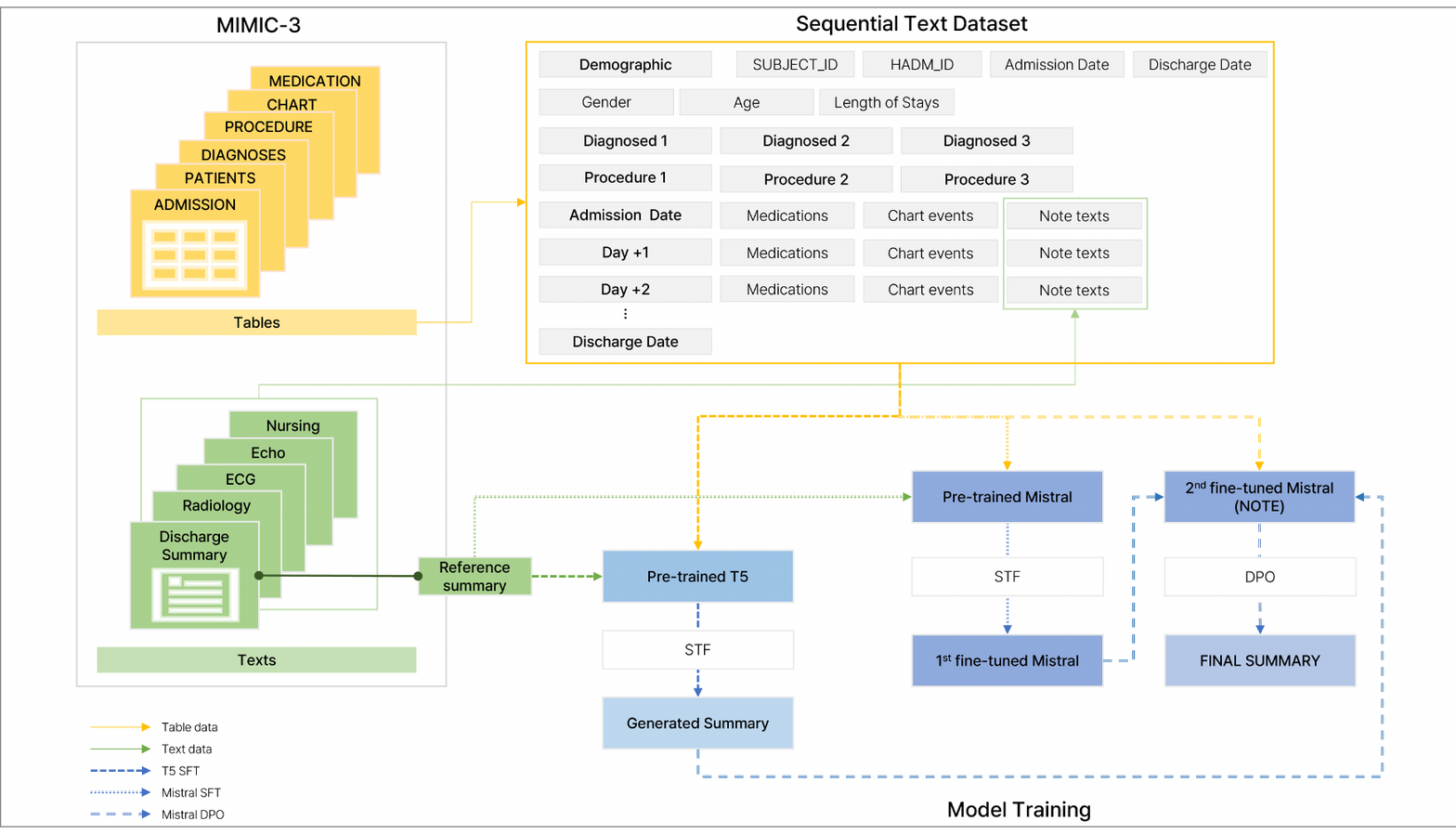}
  \caption{The overview of data processing and model training process}
  \label{fig:fig1}
\end{figure*}

Figure~\ref{fig:fig1} provides an overview of the NOTE construction process, showing the data processing and data flow required for model training. Detailed explanations are presented below.

\subsection{Preparation of dataset}
\subsubsection{Data processing}
First, we processed the table data by extracting patient demographic information and creating a data cohort. Our cohort consisted of adults aged 19 years or older who had been hospitalized for less than 7 days. This selection criterion was used because MIMIC-III consists primarily of ICU patients and because of the length limitations of the baseline model. 

We then extracted DS from the NOTEEVENTS table, which contains data from 15 categories detailed in Table ~\ref{tab:table1}. We systematically removed repeated special characters and de-identified words such as last name, physician name, etc. In addition, for comparison to the baseline model, we focused on patients with summaries no longer than 500 words. This process resulted in a cohort of 709 hospitalizations.

\begin{table}
\caption{The number of notes by category in NOTEEVENTS}
%\tiny
\centering

\label{tab:table1}
\begin{tabular}{lc}
\toprule
Category &The number of notes \\ 
\midrule
Nursing/other & 822,497 \\
Radiology & 522,279 \\
Nursing & 223,556 \\
ECG & 209,051 \\
Physician & 141,624 \\
DS & 59,652 \\
Echo & 45,794 \\
Respiratory & 31,739 \\
Nutrition & 9,418 \\
General & 8,301 \\
Rehab Services & 5,431 \\
Social Work & 2,670 \\
Case Management & 967 \\
Pharmacy & 103 \\
Consult & 98 \\
\bottomrule
\end{tabular}
\end{table}

The names of diagnoses and procedures primarily serve as records for insurance processing, while the precise prescription dates were not specified, so we categorized them as demographic information. All codes are recorded in the ICD-9\textsuperscript{th} version format, with up to 5 codes per hospitalization.

Within the MEDICATIONS table, which encompasses all drugs administered during hospitalization, we exclusively utilized the list of drugs prescribed for more than 10\% of all patients. Additionally, we factored in the duration of drug prescriptions to gauge the severity of the condition for which the drug was prescribed.

CHARTEVENTS contains various data, including laboratory test results, vital sign measurements, and physical information. LABEVENTS, a subset of CHARTEVENTS, was not utilized in our analysis. Similar to our approach with MEDICATIONS, we extracted chart data for items tested on more than 50\% of patients to prevent excessive input data length.
Furthermore, we extracted additional text data stored in the NOTEEVENTS table, representing the patient's condition and progress. This included the following list of items:
\begin{itemize}
\item Radiology Reports (2,177 cases): Extracted from X-rays, CT scans, MRIs, and various radiological tests, focusing on "FINAL REPORT" sections containing impressions and findings. 
\item Nursing Notes (126 cases): Reflect nursing care and patient progress, covering Response, Assessment, Action, Plan, and related details.
\item Nursing/Other notes (2,802 cases): Varied content from caregivers beyond nursing notes.
\item ECG Reports (791 cases): Brief Electrocardiogram reports typically with 2-3 sentences.
\item Physician Notes (88 cases): Clinician-written summaries of patient progress.
\item Echo Reports (91 cases): Extracted from Echoencephalography reports beyond the "Conclusions:" section.
\end{itemize}
\subsubsection{Sequential dataset}
We rearranged all extracted data based on the length of hospitalization. For each patient, the admission date was placed in the first row, the discharge date in the last row, and events were sorted chronologically. The tables and text data were sorted accordingly, as depicted in Figure~\ref{fig:fig1}.
\subsubsection{Training dataset}
We created three training datasets: one utilizing only table data, another utilizing only text data, and a third incorporating both table and text data. 
Although the amount of text data was relatively small compared to the table data, we added it to evaluate the performance of the model when trained on text data only. 
In addition, a dataset with a combination of table and text was created to evaluate the model's ability to understand and generate summaries from mixed inputs. This data was used to develop both the baseline model and NOTE. 

The dataset comprises a total of 709 rows, with a split ratio of 80\% for the training set (including validation) and 20\% for the test set. This results in 453 rows for training (including validation), 114 rows for validation, and 142 rows for testing.

\subsection{Experiment}
\subsubsection{Model development}
We utilized T5\cite{t5} as our baseline model, and the foundational model for NOTE is built on Mistral-7B-Instruct-v0.2(Mistral-7B)\cite{MistralAI}. Also, we utilized packages for experiment including trl\cite{trl}, transformers\cite{transformers}, and Hugging Face\cite{HuggingFace} for supervised fine-tuning (SFT) and DPO. Our experiments can be broadly divided into two steps.

Firstly, we utilized the base model T5, conducting SFT with three datasets and the DS of NOTETABLES as the reference (label). 
The T5 trained on both table and text datasets is referred to as SFT-T5. The summaries generated by this model were used in subsequent steps.

Secondly, we performed the both of SFT and DPO\cite{DPO} training using the Mistral-7B. 
In this process, we first performed SFT using the same input data and reference that were used for training SFT-T5. Additionally, QLoRA\cite{qlora}, one of Parameter-Efficient Fine Tuning (PEFT) \cite{peft} was employed in conjunction. PEFT is a highly efficient technique that freezes most of the parameters of LLM and fine-tunes only a few. This approach maintains the performance of existing LLM models while significantly reducing computational resources, such as training time and memory. It prevents catastrophic forgetting and enables efficient learning. Two popular PEFT methods are LoRA\cite{lora} and QLoRA. LoRA adds layers (adapters) to the existing LLM, allowing it to adapt to downstream tasks faster than fine-tuning the entire LLM. QLoRA provides higher efficiency than LoRA by loading quantized 4-bit weights into memory. 

Subsequently, we employed DPO, which is one of the optimization algorithms, developed based on reinforcement learning from human validation (RLHF)\cite{rlhf}. 
The current RLHF method necessitates LLM pre-training and human data annotation, followed by reinforcement learning with the reward model. In contrast, DPO utilizes preferred dataset and rejected dataset directly, eliminating the need for annotation and the reward model. This approach reduces learning instability and enables efficient optimization. Although it has the disadvantage of requiring a preferred and rejected dataset, we used this dataset as a reference (e.g., DS in NOTETABLES) and summaries generated by SFT-T5, respectively. 

Consequently, NOTE was built by adding instructions to Mistral-7B for which DPO learning was completed. Instructions contain information from existing references, such as patient information, diagnosis information, medication details, test results, summary of notes.

\subsubsection{Evaluation}
For quantitative evaluation, we have selected several metrics commonly used in NLP tasks\cite{nlp_metrics}, listed below along with brief explanations: 
\begin{itemize}
\item Mean Modified Logarithm Units (MMLU): a custom metric  developed using ChatGPT-4\cite{openai_chatgpt} that quantifies the difference in LLMs predicted probabilities between reference and generated sentences. It's a unit used in information theory\cite{mmlu}. It is derived by running the model on both inputs to obtain the logit, then applying log softmax to calculate the log probability. The average difference in log odds between the reference and generated sentences constitutes the MMLU score, the magnitude of which indicates the degree of difference between the two sentences. We then utilized BERT, T5, and Mistral-7B to calculate the MMLU for each. The Algorithm~\ref{al:al1} is a pseudo code that calculates MMLU.

\begin{equation}
MMLU = -\frac{1}{n} \sum_{i=1}^{n} \log_e(p_i).
\end{equation}

\item Recall-Oriented Understudy for Gisting Evaluation (ROUGE)\cite{rouge}: Measured based on N-grams or Longest Common Subsequence, higher scores indicate greater inclusion of common words or phrases.
\item Bilingual Evaluation Understudy (BLEU)\cite{blue1, blue2}: Similar to ROUGE, BLEU evaluates lexical similarity using N-grams, with higher scores reflecting increased lexical overlap.
\item Bidirectional Encoder Representations from Transformers (BERT) score\cite{bertscore}:Utilizing BERT\cite{bert} embeddings, this metric measures contextual information at the token level, with higher scores indicating greater similarity between two sentences.
\item Perplexity\cite{perplexity}: A measure of how well data fits the model, indicating the model's ability to accurately predict the next word based on context. Lower values signify the model's ability to predict the next word accurately in the context of meaning. Mistral-7B was employed for this metric.
\item Metric for Evaluation of Translation with Explicit Ordering (METEOR)\cite{meteor}: Primarily used in translation tasks, METEOR considers word-to-word, matching, and order and structure, with higher scores indicating more natural translations.
\end{itemize}

\begin{algorithm}
\caption{MMLU Calculation using LLM}
\label{al:al1}
\begin{algorithmic}[1]
\Require $R$ (reference sentences), $P$ (predicted sentences), $model\_path$ (LLM file path)
\Ensure $mmlu$ (Mean Modified Logarithm Units)
\State $T_R \gets \Call{Tokenize}{R}$
\State $T_P \gets \Call{Tokenize}{P}$
\State $L_R \gets \Call{LLM}{T_R, model\_path}$
\State $L_P \gets \Call{LLM}{T_P, model\_path}$
\State $min\_len \gets \min(\Call{Length}{L_R}, \Call{Length}{L_P})$
\State $L_R' \gets L_R[:, :min\_len, :]$
\State $L_P' \gets L_P[:, :min\_len, :]$
\State $mmlu\_score \gets \frac{1}{N} \sum (L_R' - L_P')$ \Comment{where $N$ is the number of elements in logits tensors}
\State \Return $mmlu\_score$
\end{algorithmic}
\end{algorithm}

For qualitative evaluation, we utilized the GPT4 Assistant API ("gpt-4-1106-preview") \cite{openai_assistant}and established the following evaluation criteria to assess the quality of generated summaries:
\begin{itemize}
\item Accuracy: Measures how accurately the summary reflects the main content and contextual meaning of the input data.
\item (Information) Retention: Assesses the ability of the summary to retain key information and details from the input data.
\item Objectivity: Evaluates whether the summary maintains the objectivity of the input data, checking for the addition of information not present in the original that could distort the input data.
\item Structure: Examines the systematic organization and orderly arrangement of the summary.
\item Coherence: Focuses on the logical and consistent semantic structure of the summary, especially the connections between sentences.
\item Grammar: Checks the grammatical accuracy of the summary, including correct grammar, expressions, and punctuation.
\item Readability: Considers the clarity and readability of the summary, ensuring that it effectively conveys the content without unnecessary repetition or verbosity.
\end{itemize}

Ten DSs from the test set were randomly selected and evaluated using the GPT4 Assistant API. The sequential input dataset was provided as the first input, followed by the summaries generated by the SFT-T5 and NOTE, respectively. GPT4 then assigned each summary a score based on the seven evaluation criteria, with each criterion scored out of 10 points.

\subsubsection{Demo Application}
Finally, we developed a demo web-based application using Gradio\cite{gradio} and Hugging Face to demonstrate the practical use of NOTE. Due to restrictions on access to MIMIC-III data for unauthorized users, we generated example fake data that resembles MIMIC-III. The application is available on Hugging Face space(\href{https://huggingface.co/spaces/jinee/note-demo}{link})

\subsubsection{Environment}
The model was developed using Python (3.8.10), the PyTorch deep learning framework (2.0.1+cu118), and Transformers (4.35.2). Model training and evaluation were carried out on an NVIDIA GeForce RTX 3090 GPU. The experiments were conducted on Ubuntu 20.04 LTS. Additional libraries utilized for data manipulation and analysis include Hugging Face Hub (0.19.4), NumPy (1.23.5), and Pandas (1.2.4).

\section{Results}
\subsection{Quantitative evaluation}

\begin{figure}
  \centering
  \includegraphics[width=\linewidth]{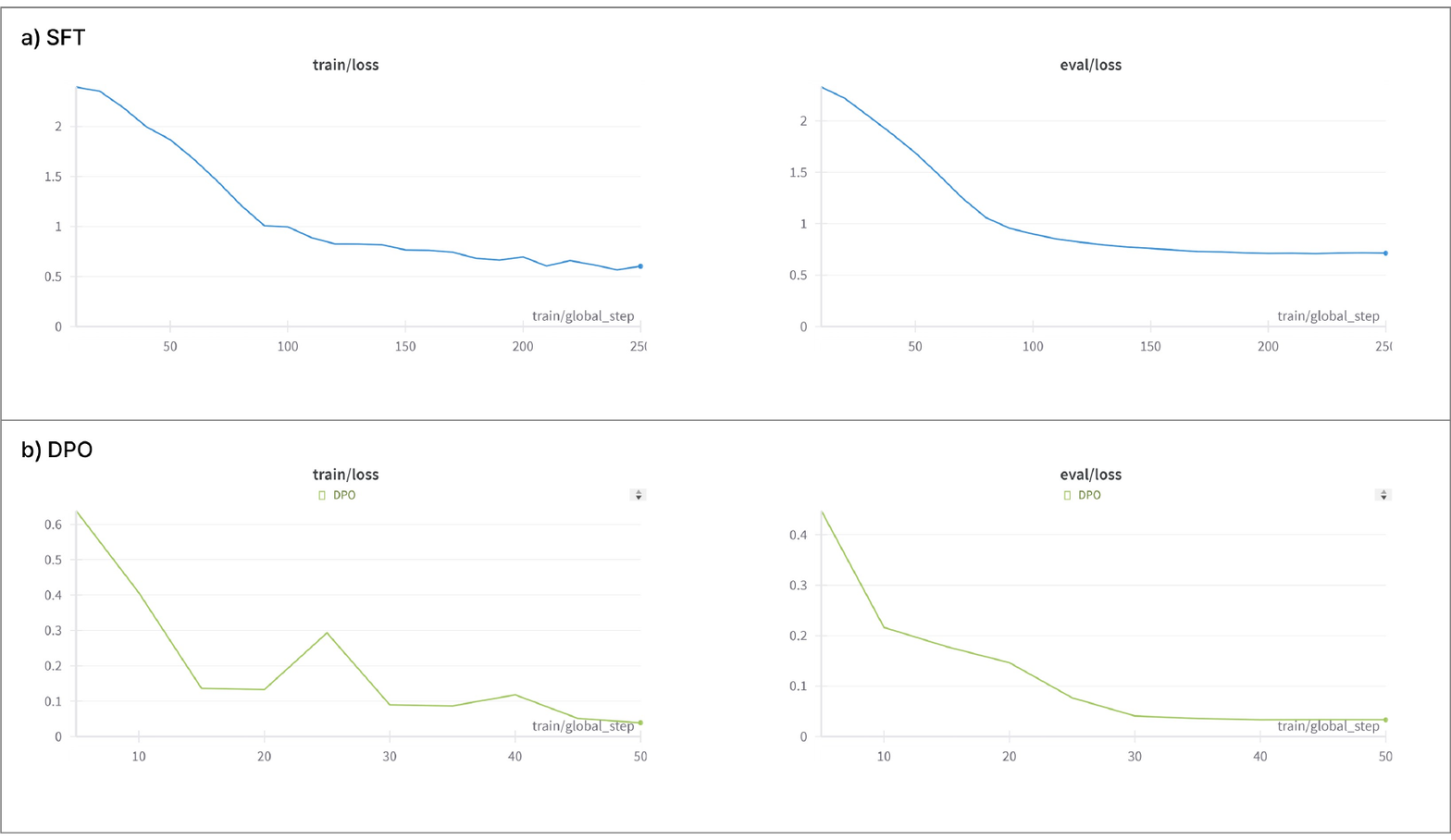}
  \caption{Training and validation loss chart in SFT(a), DPO(b) in NOTE}
  \label{fig:fig2}
\end{figure}
The training loss of the SFT is 0.60 and the validation loss is 0.71. DPO's training loss is 0.039 and validation loss is 0.033, both of which shown in Figure~\ref{fig:fig2}. Additionally, SFT was trained for 50 global steps (8 epochs, runtime: 237 minutes), and DPO was trained for 50 global steps (0.44 epochs, runtime: 71 minutes). 

\begin{table}
\centering
\caption{list of parameters for SFT and DPO with Mistral-7B}
\scriptsize
\label{tab:params}
\begin{tabular}{llcc}
\toprule
 Category & Parameter  & Value (SFT) & Value (DPO) \\ 
\midrule
LoRA & r & 16 & 16   \\
     & lora alpha & 16 & 16    \\
     & lora dropout & 0.05 & 0.05  \\
     & target & q, k, v, o, gate & q, k, v, o, gate \\
\midrule
Training& per device train batch size & 4 & 1    \\
        & per device eval batch size  & default (8)  & 1 \\
        & optimizer &   paged adamw 8bit & paged adamw 8bit \\
        & lr scheduler  & cosine  & cosine \\
        & gradient accumulation step & 2 & 2 \\

\bottomrule
\end{tabular}
\end{table}

A list of the primary parameters tuned during training can be found in the Table~\ref{tab:params}, and additional parameters can be found in Hugging Face model(\href{https://huggingface.co/jinee/note}{link})

\begin{table*}
\centering
\caption{Model performance by dataset type and PT, SFT, and DPO (The top score for each metric is highlighted in bold and the second highest score is underlined.)}
\label{tab:table2}
\scriptsize
\begin{tabular}{@{}ccccccccccccc@{}}
\toprule
\multirow{2}{*}{Dataset (N)} & \multirow{2}{*}{Model} & \multirow{2}{*}{Training} & \multicolumn{3}{c}{MMLU} & \multicolumn{3}{c}{ROUGE} & \multirow{2}{*}{BLEU} & \multirow{2}{*}{BERTscore} & \multirow{2}{*}{Perplexity} & \multirow{2}{*}{METEOR} \\ 
\cmidrule(lr){4-6} \cmidrule(lr){7-9}
 &  &  & BERT & T5 & Mistral-7B &Rouge1 & Rouge2 & RougeL &  &  &  &   \\ 
\midrule
\multirow{2}{*}{Table} & T5 & PT & -0.66 & 0.5 & -0.08 & 0.06 & 0.01 & 0.04 & 0.0 & 0.73 & 3.0e+10 & 0.03  \\
                       & T5 & SFT & -0.13 & 0.82 & 0.14 & \textbf{0.37} & \textbf{0.15} & \textbf{0.21} & \textbf{0.13} & \textbf{0.81} & 2.2e+11 & \textbf{0.23}  \\
\multirow{2}{*}{Text}  & T5 & PT & \underline{-0.05} & 0.77 & -1.44 & 0.06 & 0.01 & 0.04  & 0.0& 0.72 &  \textbf{2.9e+08} & 0.03 \\
                       & T5 & SFT & -0.14 &  \textbf{0.28} & \underline{-0.05} &  \textbf{0.37} &  \textbf{0.15} &0.2 &  \textbf{0.13} &  \textbf{0.81} & 6.5e+09 & 0.22  \\
\multirow{5}{*}{Table \& Text} & T5 & PT & -0.56 & 0.34 & -0.38 & 0.06 & 0.01 & 0.04 & 0.0 & 0.73 & 4.9e+10 & 0.03  \\
 & T5 & SFT & -0.18 & 0.63 & \textbf{-0.01} &  \textbf{0.37} &  \textbf{0.15} &  \textbf{0.21} &  \textbf{0.13} &  \textbf{0.81} & 1.9e+11 &  \textbf{0.23}  \\
 & Mistral-7B & PT & -0.07 & \underline{0.32} & 1.93 & 0.23 & 0.04  &0.11 & 0.02 & 0.76 & 2.12e+17 & 0.15 \\
 & Mistral-7B & SFT & -0.25 & 3.5 & 3.01 & 0.23  & \underline{0.05} & 0.11 & \underline{0.04} & 0.76 & 1.02e+17 & 0.15 \\
 & Mistral-7B & DPO (Ours) &  \textbf{-0.02} & 0.76 & 2.26 & \underline{0.26} &  \underline{0.05}& \underline{0.12} & 0.02 & \underline{0.77} & \underline{1.25e+13} & \underline{0.18} \\
\bottomrule
\end{tabular}
\end{table*}

Table ~\ref{tab:table2} shows the performance results of the models used to construct NOTE. The data set consists of table, text, table and text, and the model consists of T5 and Mistral-7B. The highest score for each metric is highlighted in bold and the second highest score is underlined.

We observed that the SFT-T5 scores the highest in all metrics except for MMLU. Particularly, the SFT-T5 trained on datasets using only tables or both tables and text achieved high ROUGE, BLEU, and BERTscore. Despite this, NOTE secured the second-highest scores in most metrics, with only about a 0.04 difference in BERTscore. This suggests that the sentences generated by SFT-T5 have a similar sentence structure to the reference sentences. However, as confirmed in the subsequent qualitative evaluation, generated summary by SFT-T5 often contained errors in values.

\subsection{Qualitative evaluation}

\begin{table*}
\caption{Qualitative evaluation of generated summaries by SFT-T5 and NOTE}
%\tiny

\label{tab:table3}
\begin{tabular}{lccccccccc}
\toprule
 & Accuracy & Retention & Objectivity & Structure & Coherence & Grammar  & Readability & Total \\ 
\midrule
SFT-T5 & 3.9 & 4.1 & 4.6 & 4.2 & 4 & 5.3 & 4.1 & 30.2 \\
 & ($\pm$1.3) & ($\pm$1.14) & ($\pm$1.8) & ($\pm$0.87) & ($\pm$1.09) & ($\pm$1.0) & ($\pm$1.04) & ($\pm$7.75) \\
\addlinespace
NOTE & 8.3 & 8.2 & 9.1 & 8.7 & 8.1 & 8.7 & 8.2 & 59.3 \\
 & ($\pm$0.64) & $\pm$(0.87) & ($\pm$0.83) & ($\pm$0.45) & ($\pm$0.7) & ($\pm$0.46) & ($\pm$0.24) & ($\pm$3.03) \\
\bottomrule
\end{tabular}
\end{table*}

In this section, we qualitatively evaluated the summaries generated by SFT-T5 and NOTE. Ten people were randomly selected from the test set, and the perfect score for each evaluation indicator is 10 points, with a total score of 70 points. As shown in Table ~\ref{tab:table3}, NOTE obtained high scores in all evaluation indicators, and it was confirmed that it achieved twice the performance of SFT-T5 in Information retention, which evaluates whether content that is not in the input data was written. 

The example of generated summaries and evaluation results can be found in the following table ~\ref{tab:table4}. In this case, SFT-T5 correctly generated the category name "Discharge Medications:", but all subsequent drug names are not present in the input data, indicating repetitive generation. Conversely, it was observed that the drugs generated by NOTE correctly distinguished between those prescribed during hospitalization and those prescribed at discharge. 

\begin{table*}
%\scriptsize
\tiny
\caption{SFT-T5 and NOTE generated summaries and GPT4 assessment results for 1 randomly selected patient (The highlighted sections indicate errors or were identified in the evaluation.)}
\label{tab:table4}
\begin{tabular}{lp{9cm}p{6cm}}
\toprule
Model & Generated summary & Evaluation by GPT4 Assistant API \\
\midrule

SFT-T5 & admission date:2205-2-9 discharge date: 2205-2-31 date of birth:2040-3-16 sex: m service: history of present illness: \newline the patient is a 80-year-old gentleman with a history of gastroesophageal laceration and hemorrhage. he had a past cal history of atrial fibrillation. \newline past surgical history: 1. gastrostomy.\hl{2. hypertension. 3. hypertension} in the upper extremity. 3. gastrostritis. 3. hepatitis \hl{b.i.d.} \newline \hl{allergies: penicillin.} \newline physical examination: on physical examination, the patient was afebrile. lungs were clear to auscultation bilaterally. abdomen was soft, nontender, nondistended. extremities were warm and well perfused.  \newline hospital course: he was admitted to the surgical intensive care unit where he underwent \hl{a subarachnoid gastric bypass bypass graft times three.} the patient underwent an endoscopic percutaneous stenosis of the lower extremity and was transferred to the csru. his vital signs were stable. his hematocrit was regular rate and rhythm with no murmurs, rubs or gallops. his foley catheter was discontinued on postoperative day number one. his blood pressure was 109/60. \hl{his inr was 98.9}, and the patient tolerated the procedure well and was discharged home in stable condition. \newline discharge cations: \hl{1. aspirin 325 mg p.o. q. day. 2. colace 100 mg once daily. 2. sulfa 100 mg twice a day. 3. colace 50 mg q day. 4. tylenol 50 mg once per day. 5. systolic acid 100 mg tid. 4. colace 500 mg once a week. 6. heparin 10 mg once weekly. 7. colace 75 mg once every two days. 7. tachycardia. 8. hepatoma. 9. hepatol 25 mg ppd. 6. colace 20 mg once monthly. 7. percocet one to two tablets po q h.s. 6. aspirine 100 mg po bid. 8. colace 200 mg bifib. 7. as} 

& 1) Accuracy: *2/10*.\newline The summary includes incorrect discharge date formatting, lists nonexistent surgical histories such as "subarachnoid gastric bypass bypass graft times three," and includes incorrect medications and diagnoses that are not in line with the input data. \newline \newline
2) Retention: *1/10*. \newline Shows poor retention of the patient's actual history and procedures, with many added inaccuracies that are not present in the input data.\newline \newline
3) Objectivity: *2/10*. \newline Introduces details not present in the input data, significantly reducing objectivity.\newline \newline
4) Structure: *3/10*. \newline The structure is poor, with a confusing arrangement of medical terms and treatment descriptions.\newline \newline
5) Coherence: *2/10*.\newline Incoherent due to the inclusion of many erroneous details, with the narrative disrupted by nonsensical phrases.\newline \newline
6) Grammar: *3/10*.\newline Contains numerous grammatical errors and inappropriate use of medical terminology.\newline \newline
7) Readability: *2/10*.\newline Difficult to read due to numerous inaccuracies, confusing language, and poor structure.\newline  \newline Summary of Scores: *15/70*  \\
\midrule

NOTE & 1. Patient information:\newline 
- SUBJECT\_ID: 946 \newline 
- HADM\_ID: 149258.0 \newline 
- Hospitalization date: 2021-05-29\newline 
- Discharge date: 2021-05-31\newline 
- Length of stay: 2 days\newline 
- Gender: Male\newline 
- Date of birth: 2040-03-16\newline 
- Age: 80\newline 
- Allergy: V551 Attention to gastrostomy\newline \newline 
2. Diagnostic information and past history:\newline 
- Atrial fibrillation (D1)\newline 
- Gastroesophageal laceration-hemorrhage syndrome (D2)\newline 
- Acute venous embolism and thrombosis of unspecified deep vessels of lower extremity (D3)\newline 
- Congestive heart failure, unspecified (D4)\newline 
- Percutaneous endoscopic gastrostomy (PEG) (P0)\newline \newline 
3. Surgery or procedure information:\newline 
- PEG procedure done on 2021-05-30\newline \newline 
4. Significant medication administration during hospitalization and discharge medication history:\newline 
- Heparin Sodium: Administered intravenously twice on 2021-05-29 and once on 2021-05-30\newline 
- Levofloxacin: Administered intravenously on 2021-05-30\newline 
- Metoprolol: Administered intravenously five times and orally twice\newline 
- Fentanyl Citrate: Administered intravenously twice\newline 
- Propofol: Administered intravenously twice\newline 
- Magnesium Sulfate: Administered intravenously twice\newline 
- Warfarin: Discharge medication\newline 
- Lorazepam: Discharge medication\newline 
- Haloperidol: Discharge medication\newline \newline 
5. Meaningful lab tests during hospitalization:\newline 
- Creatinine: 0.6 and 0.7\newline 
- Platelets: 222.0, 187.0, 171.0\newline 
- Hematocrit: 30.5, 30.0, 29.2\newline 
- Hemoglobin: 10.2, 10.3, 9.9\newline 
- INR: 1.2 and 1.3\newline 
- PT/INR: 26.4 and 29.6\newline \newline 
6. Summary of significant text records/notes:\newline 
- On 2021-05-30, the patient underwent a PEG procedure, which was tolerated well. The patient was weaned off the ventilator and started on cpap +7.5 and 18ips. Heparin was to be restarted that evening.\newline 
- On 2021-05-31, the patient's atrial fibrillation became rapid, and they exhibited agitation. Haloperidol and lorazepam were given without effect. Cardizem was administered, and the heart rate decreased below 100. The patient's PEG tube site was checked, and no bleeding or drainage was observed. The team was instructed to begin tube feedings and prepare for transfer to rehab. The respiratory care team reported no changes.\newline \newline 
7. Discharge outcomes and treatment plan:\newline 
- The patient was transferred to rehab on 2021-05-31.\newline 
- Warfarin, lorazepam, and haloperidol were prescribed as discharge medications.\newline \newline 
8. Overall summary:\newline 
The patient, an 80-year-old male, was hospitalized from 2021-05-29 to 2021-05-31 due to various conditions, including atrial fibrillation, gastroesophageal laceration-hemorrhage syndrome, acute venous embolism and thrombosis, and congestive heart failure. During his hospitalization, he underwent a PEG procedure, which was well-tolerated. He received multiple medications, including heparin, levofloxacin, metoprolol, fentanyl, and propofol.
 & 1) Accuracy:*8/10*. \newline Remains consistent with the diagnoses and procedural information provided in the input data, with small errors related to the specifics of medication administration not mentioned in the input.\newline \newline
2) Retention:*7/10*. \newline Adequately retains important details such as the medical condition, treatment received, and medications administered, despite minor omissions.\newline \newline
3) Objectivity:*8/10*. \newline Mostly maintains objectivity by sticking to the facts and avoiding unfounded additions.\newline \newline 
4) Structure:*9/10*.\newline  Presents a clear and systematic structure, making it easy to understand the patient's clinical timeline.\newline \newline
5) Coherence:*8/10*. \newline The report is logically constructed and provides a coherent account of the hospital course.\newline  \newline
6) Grammar:*9/10*.\newline  Well written with few grammatical issues, contributing to the clarity of the summary.\newline \newline
7) Readability:*8/10*. \newline  Clear and understandable flow of information with good readability, helping readers grasp the patient’s situation relatively easily. \newline \newline Summary of Scores:*57/70* \\ 

\bottomrule
\end{tabular}
\end{table*}

\subsection{Demo Application}

\begin{figure}
  \centering
 \scriptsize

  \includegraphics[width=\linewidth]{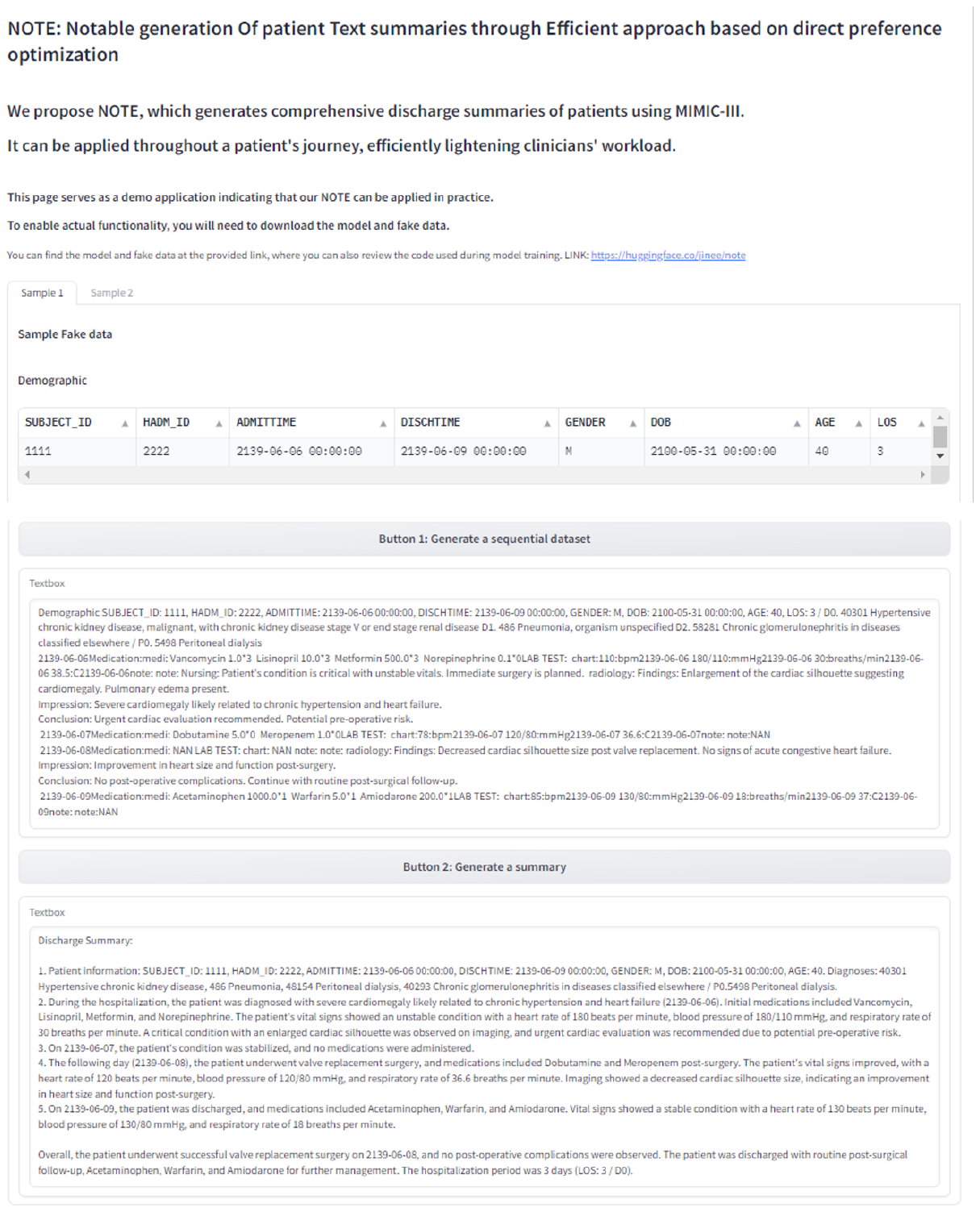}
  \caption{NOTE demonstration application}
  \label{fig:fig3}
\end{figure}

The demo application created based on NOTE is for two sample patients and presents data during a virtual hospitalization for each patient.
As shown in figure~\ref{fig:fig3}, user can create a sequential dataset by pressing button 1 (Generate a sequential dataset), and by pressing button 2 (Generate a summary), NOTE will receive this dataset as input and reproduce the process of creating a DS. Due to space limitations, actual operation requires downloading the NOTE and creating a personal environment.

\section{Conclusion}
\subsection{Applicability in medicine}
Firstly, we propose that our NOTE can seamlessly integrate tabular and textual data stored in various formats within EMRs to generate a cohesive sequential dataset. This capability can significantly alleviate the burden on medical staff who would otherwise need to manually review vast amounts of information when compiling reports.
Additionally, NOTE was created using the instruct model, which means it can be expanded into a chatbot that provides detailed information about the individual medical record.

Secondly, given the paramount importance of safeguarding patients' personal information, many medical institutions face challenges in utilizing API from open-source models. Therefore, we advocate for the utilization of NOTE in the medical domain. With NOTE, fine-tuning the model requires minimal computing resources, and adjustments to the foundation model could be made with flexibility.

Lastly, we propose that the demo application can be practically deployed by integrating it with existing EMRs within medical institutions. This integration would facilitate seamless utilization of NOTE's capabilities in real-world clinical settings.

\subsection{Limitations and future works}
Firstly, the decision to shorten the hospitalization period during the dataset creation process was due to the limited input length of T5. Additionally, the absence of image files in MIMIC-III restricted the inclusion of images in the dataset.
However, LLMs such as Mistral-7B have longer input lengths, enabling further analysis of the entire hospitalization period. Future studies using MIMIC-IV are planned to incorporate image data alongside existing text data for multi-modal research. Additionally, partnerships with institutions like Asan Medical Center are being established to generate summaries based on real data.

Secondly, the variation in generated summaries based on prompt engineering is acknowledged as a fundamental issue in generative AI. NOTE requires specific instructions to ensure consistent output, presenting opportunities to flexibly adapt existing report templates in medical institutions. Future works will also address this aspect.

Another limitation observed is the return of unidentified unrealistic dates converted to historically documented dates, despite efforts to use the dates as they are. However, accurate reflection of age and event creation dates indicates a potential solution, given that realistic dates are predominantly used in actual medical contexts.

Thirdly, most quantitative evaluation metrics for summarization tasks assess word overlap or contextual similarity. However, these metrics may not be suitable for evaluating generated summary by NOTE, as most summarization datasets have long input data and short key sentence references\cite{extractive, abstractive}. 
It is essential to develop objective metrics that can evaluate proper information integration, accurate summary generation in similar tasks.

Although a qualitative evaluation was conducted to address this limitation, it is important to note that relying solely on the GPT4 Assistant API may be seen as a constraint. Future plans include incorporating human validation to reflect feedback from clinicians.

In conclusion, NOTE represents a significant advancement in medical documentation by providing a comprehensive and efficient solution for generating discharge summaries and more. By leveraging DPO and PEFT, NOTE goes beyond mere discharge summaries to effectively cover the entire patient journey. This approach has the potential to reduce the workload of clinicians, improve the efficiency of patient care, and contribute to the development of more effective healthcare systems.

% \section*{Acknowledgments}

%Bibliography
\bibliographystyle{unsrt}  
\bibliography{references}  

\begin{thebibliography}{10}

\bibitem{patient_journey}
Timothy~M Trebble, Navjyot Hansi, Theresa Hydes, Melissa~A Smith, and Marc Baker.
\newblock Process mapping the patient journey: an introduction.
\newblock {\em Bmj}, 341, 2010.

\bibitem{emr}
Faustine Williams and Suzanne~Austin Boren.
\newblock The role of the electronic medical record (emr) in care delivery development in developing countries: a systematic review.
\newblock {\em Informatics in primary care}, 16(2), 2008.

\bibitem{discharge_important}
SF~Murphy, L~Lenihan, F~Orefuwa, G~Colohan, I~Hynes, and CG~Collins.
\newblock Electronic discharge summary and prescription: improving communication between hospital and primary care.
\newblock {\em Irish Journal of Medical Science (1971-)}, 186:455--459, 2017.

\bibitem{mimic_prediction}
Chuizheng Meng, Loc Trinh, Nan Xu, James Enouen, and Yan Liu.
\newblock Interpretability and fairness evaluation of deep learning models on mimic-iv dataset.
\newblock {\em Scientific Reports}, 12(1):7166, 2022.

\bibitem{llm_medicine}
Arun~James Thirunavukarasu, Darren Shu~Jeng Ting, Kabilan Elangovan, Laura Gutierrez, Ting~Fang Tan, and Daniel Shu~Wei Ting.
\newblock Large language models in medicine.
\newblock {\em Nature medicine}, 29(8):1930--1940, 2023.

\bibitem{llm_gpt}
Yiheng Liu, Tianle Han, Siyuan Ma, Jiayue Zhang, Yuanyuan Yang, Jiaming Tian, Hao He, Antong Li, Mengshen He, Zhengliang Liu, et~al.
\newblock Summary of chatgpt-related research and perspective towards the future of large language models.
\newblock {\em Meta-Radiology}, page 100017, 2023.

\bibitem{DPO}
Rafael Rafailov, Archit Sharma, Eric Mitchell, Stefano Ermon, Christopher~D Manning, and Chelsea Finn.
\newblock Direct preference optimization: Your language model is secretly a reward model.
\newblock {\em arXiv preprint arXiv:2305.18290}, 2023.

\bibitem{mimic_1}
Alistair Johnson, Tom Pollard, and Roger Mark.
\newblock Mimic-iii clinical database (version 1.4).
\newblock {\em PhysioNet}, 10(C2XW26):2, 2016.

\bibitem{mimic_2}
Alistair~EW Johnson, Tom~J Pollard, Lu~Shen, Li-wei~H Lehman, Mengling Feng, Mohammad Ghassemi, Benjamin Moody, Peter Szolovits, Leo Anthony~Celi, and Roger~G Mark.
\newblock Mimic-iii, a freely accessible critical care database.
\newblock {\em Scientific data}, 3(1):1--9, 2016.

\bibitem{mimic_3}
Ary~L Goldberger, Luis~AN Amaral, Leon Glass, Jeffrey~M Hausdorff, Plamen~Ch Ivanov, Roger~G Mark, Joseph~E Mietus, George~B Moody, Chung-Kang Peng, and H~Eugene Stanley.
\newblock Physiobank, physiotoolkit, and physionet: components of a new research resource for complex physiologic signals.
\newblock {\em circulation}, 101(23):e215--e220, 2000.

\bibitem{genAI}
Krishnaram Kenthapadi, Himabindu Lakkaraju, and Nazneen Rajani.
\newblock Generative ai meets responsible ai: Practical challenges and opportunities.
\newblock In {\em Proceedings of the 29th ACM SIGKDD Conference on Knowledge Discovery and Data Mining}, pages 5805--5806, 2023.

\bibitem{gemini}
{DeepMind}.
\newblock {Gemini - DeepMind Technologies}, 2024.
\newblock Accessed: "Feburary 08, 2024.

\bibitem{llama2}
Hugo Touvron, Louis Martin, Kevin Stone, Peter Albert, Amjad Almahairi, Yasmine Babaei, Nikolay Bashlykov, Soumya Batra, Prajjwal Bhargava, Shruti Bhosale, et~al.
\newblock Llama 2: Open foundation and fine-tuned chat models.
\newblock {\em arXiv preprint arXiv:2307.09288}, 2023.

\bibitem{extractive}
Ridam Srivastava, Prabhav Singh, KPS Rana, and Vineet Kumar.
\newblock A topic modeled unsupervised approach to single document extractive text summarization.
\newblock {\em Knowledge-Based Systems}, 246:108636, 2022.

\bibitem{abstractive}
Som Gupta and Sanjai~Kumar Gupta.
\newblock Abstractive summarization: An overview of the state of the art.
\newblock {\em Expert Systems with Applications}, 121:49--65, 2019.

\bibitem{t5}
Colin Raffel, Noam Shazeer, Adam Roberts, Katherine Lee, Sharan Narang, Michael Matena, Yanqi Zhou, Wei Li, and Peter~J Liu.
\newblock Exploring the limits of transfer learning with a unified text-to-text transformer.
\newblock {\em The Journal of Machine Learning Research}, 21(1):5485--5551, 2020.

\bibitem{openai_chatgpt}
{OpenAI}.
\newblock {ChatGPT-4}, 2024.
\newblock Accessed: "Feburary 08, 2024".

\bibitem{openai_assistant}
{OpenAI}.
\newblock {OpenAI Assistant Documentation}, 2024.
\newblock Accessed: "Feburary 08, 2024".

\bibitem{mimic_xgb}
Nianzong Hou, Mingzhe Li, Lu~He, Bing Xie, Lin Wang, Rumin Zhang, Yong Yu, Xiaodong Sun, Zhengsheng Pan, and Kai Wang.
\newblock Predicting 30-days mortality for mimic-iii patients with sepsis-3: a machine learning approach using xgboost.
\newblock {\em Journal of translational medicine}, 18(1):1--14, 2020.

\bibitem{mimic_nature}
Stephanie~L Hyland, Martin Faltys, Matthias H{\"u}ser, Xinrui Lyu, Thomas Gumbsch, Crist{\'o}bal Esteban, Christian Bock, Max Horn, Michael Moor, Bastian Rieck, et~al.
\newblock Early prediction of circulatory failure in the intensive care unit using machine learning.
\newblock {\em Nature medicine}, 26(3):364--373, 2020.

\bibitem{sum_progress}
Yanjun Gao, Timothy Miller, Dongfang Xu, Dmitriy Dligach, Matthew~M Churpek, and Majid Afshar.
\newblock Summarizing patients’ problems from hospital progress notes using pre-trained sequence-to-sequence models.
\newblock In {\em Proceedings of COLING. International Conference on Computational Linguistics}, volume 2022, page 2979. NIH Public Access, 2022.

\bibitem{medpalm_google}
Karan Singhal, Shekoofeh Azizi, Tao Tu, S~Sara Mahdavi, Jason Wei, Hyung~Won Chung, Nathan Scales, Ajay Tanwani, Heather Cole-Lewis, Stephen Pfohl, et~al.
\newblock Large language models encode clinical knowledge.
\newblock {\em Nature}, 620(7972):172--180, 2023.

\bibitem{gatortron_nvidia}
Xi~Yang, Aokun Chen, Nima PourNejatian, Hoo~Chang Shin, Kaleb~E Smith, Christopher Parisien, Colin Compas, Cheryl Martin, Mona~G Flores, Ying Zhang, et~al.
\newblock Gatortron: A large clinical language model to unlock patient information from unstructured electronic health records.
\newblock {\em arXiv preprint arXiv:2203.03540}, 2022.

\bibitem{small_llm}
Aakanksha Chowdhery, Sharan Narang, Jacob Devlin, Maarten Bosma, Gaurav Mishra, Adam Roberts, Paul Barham, Hyung~Won Chung, Charles Sutton, Sebastian Gehrmann, et~al.
\newblock Palm: Scaling language modeling with pathways.
\newblock {\em Journal of Machine Learning Research}, 24(240):1--113, 2023.

\bibitem{peft}
Sourab Mangrulkar, Sylvain Gugger, Lysandre Debut, Younes Belkada, Sayak Paul, and Benjamin Bossan.
\newblock Peft: State-of-the-art parameter-efficient fine-tuning methods.
\newblock \url{https://github.com/huggingface/peft}, 2022.

\bibitem{MistralAI}
{Mistral AI}, 2024.
\newblock Accessed: "Feburary 08, 2024".

\bibitem{trl}
Leandro von Werra, Younes Belkada, Lewis Tunstall, Edward Beeching, Tristan Thrush, Nathan Lambert, and Shengyi Huang.
\newblock Trl: Transformer reinforcement learning.
\newblock \url{https://github.com/huggingface/trl}, 2020.

\bibitem{transformers}
Thomas Wolf, Lysandre Debut, Victor Sanh, Julien Chaumond, Clement Delangue, Anthony Moi, Pierric Cistac, Tim Rault, Rémi Louf, Morgan Funtowicz, Joe Davison, Sam Shleifer, Patrick von Platen, Clara Ma, Yacine Jernite, Julien Plu, Canwen Xu, Teven~Le Scao, Sylvain Gugger, Mariama Drame, Quentin Lhoest, and Alexander~M. Rush.
\newblock Transformers: State-of-the-art natural language processing.
\newblock In {\em Proceedings of the 2020 Conference on Empirical Methods in Natural Language Processing: System Demonstrations}, pages 38--45, Online, October 2020. Association for Computational Linguistics.

\bibitem{HuggingFace}
{Hugging Face}, 2024.
\newblock Accessed: "Feburary 08, 2024".

\bibitem{qlora}
Tim Dettmers, Artidoro Pagnoni, Ari Holtzman, and Luke Zettlemoyer.
\newblock Qlora: Efficient finetuning of quantized llms.
\newblock {\em arXiv preprint arXiv:2305.14314}, 2023.

\bibitem{lora}
Edward~J Hu, Yelong Shen, Phillip Wallis, Zeyuan Allen-Zhu, Yuanzhi Li, Shean Wang, Lu~Wang, and Weizhu Chen.
\newblock Lora: Low-rank adaptation of large language models.
\newblock {\em arXiv preprint arXiv:2106.09685}, 2021.

\bibitem{rlhf}
Paul~F Christiano, Jan Leike, Tom Brown, Miljan Martic, Shane Legg, and Dario Amodei.
\newblock Deep reinforcement learning from human preferences.
\newblock {\em Advances in neural information processing systems}, 30, 2017.

\bibitem{nlp_metrics}
Nouf~Ibrahim Altmami and Mohamed El~Bachir Menai.
\newblock Automatic summarization of scientific articles: A survey.
\newblock {\em Journal of King Saud University-Computer and Information Sciences}, 34(4):1011--1028, 2022.

\bibitem{mmlu}
Claude~Elwood Shannon.
\newblock A mathematical theory of communication.
\newblock {\em The Bell system technical journal}, 27(3):379--423, 1948.

\bibitem{rouge}
Chin-Yew Lin.
\newblock {ROUGE}: A package for automatic evaluation of summaries.
\newblock In {\em Text Summarization Branches Out}, pages 74--81, Barcelona, Spain, July 2004. Association for Computational Linguistics.

\bibitem{blue1}
Kishore Papineni, Salim Roukos, Todd Ward, and Wei jing Zhu.
\newblock Bleu: a method for automatic evaluation of machine translation.
\newblock pages 311--318, 2002.

\bibitem{blue2}
Chin-Yew Lin and Franz~Josef Och.
\newblock {ORANGE}: a method for evaluating automatic evaluation metrics for machine translation.
\newblock In {\em {COLING} 2004: Proceedings of the 20th International Conference on Computational Linguistics}, pages 501--507, Geneva, Switzerland, aug 23{--}aug 27 2004. COLING.

\bibitem{bertscore}
Tianyi Zhang*, Varsha Kishore*, Felix Wu*, Kilian~Q. Weinberger, and Yoav Artzi.
\newblock Bertscore: Evaluating text generation with bert.
\newblock In {\em International Conference on Learning Representations}, 2020.

\bibitem{bert}
Jacob Devlin, Ming-Wei Chang, Kenton Lee, and Kristina Toutanova.
\newblock Bert: Pre-training of deep bidirectional transformers for language understanding.
\newblock {\em arXiv preprint arXiv:1810.04805}, 2018.

\bibitem{perplexity}
Fred Jelinek, Robert~L Mercer, Lalit~R Bahl, and James~K Baker.
\newblock Perplexity—a measure of the difficulty of speech recognition tasks.
\newblock {\em The Journal of the Acoustical Society of America}, 62(S1):S63--S63, 1977.

\bibitem{meteor}
Satanjeev Banerjee and Alon Lavie.
\newblock {METEOR}: An automatic metric for {MT} evaluation with improved correlation with human judgments.
\newblock In {\em Proceedings of the {ACL} Workshop on Intrinsic and Extrinsic Evaluation Measures for Machine Translation and/or Summarization}, pages 65--72, Ann Arbor, Michigan, June 2005. Association for Computational Linguistics.

\bibitem{gradio}
Abubakar Abid, Ali Abdalla, Ali Abid, Dawood Khan, Abdulrahman Alfozan, and James Zou.
\newblock Gradio: Hassle-free sharing and testing of ml models in the wild.
\newblock {\em arXiv preprint arXiv:1906.02569}, 2019.

\end{thebibliography}

\end{document}